\documentclass[conference]{IEEEtran}
\IEEEoverridecommandlockouts
\usepackage{cite}
\usepackage{amsmath,amssymb,amsfonts}
\usepackage{algorithmic}
\usepackage{graphicx}
\usepackage{textcomp}
\usepackage{xcolor}
\usepackage{adjustbox}
\def\BibTeX{{\rm B\kern-.05em{\sc i\kern-.025em b}\kern-.08em
    T\kern-.1667em\lower.7ex\hbox{E}\kern-.125emX}}
\begin{document}

\title{Towards a Skeleton-Based Action Recognition For Realistic Scenarios\\
{}
\thanks{}
}

\author{\IEEEauthorblockN{1\textsuperscript{st} \c{C}a\u{g}atay Odaba\c{s}{\i}}
\IEEEauthorblockA{\textit{Fraunhofer IPA} \\
Stuttgart, Germany \\
cagatay.odabasi@ipa.fraunhofer.de}
\and
\IEEEauthorblockN{2\textsuperscript{nd} Jewel Jose}
\IEEEauthorblockA{\textit{University of Technology Chemnitz} \\
Regensburg, Germany \\
jewelvjose@gmail.com}

}

\maketitle

\begin{abstract}
Understanding human actions is a crucial problem for service robots. However, the general trend in Action Recognition is developing and testing these systems on structured datasets. That's why this work presents a practical Skeleton-based Action Recognition framework which can be used in realistic scenarios. Our results show that although non-augmented and non-normalized data may yield comparable results on the test split of the dataset, it is far from being useful on another dataset which is a manually collected data. 
\end{abstract}


\begin{figure}[h!]
	\centering
	\includegraphics[width=0.95\linewidth]{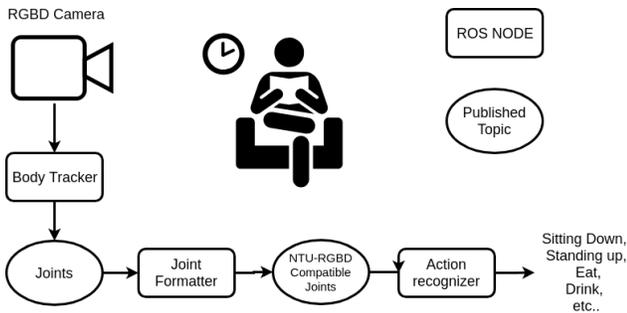}
	\caption{The flowchart of the real-world Implementation of Our Action Recognition ROS(Robot Operating System) Package.}
	\label{fig:liveaction}
\end{figure}

\section{Introduction}

The service robots need to share the environments with people such as hospitals or care-houses. Therefore, the robots should be aware of where the people are and what they are doing by using the sensory data. However, how transferable are the action recognition systems?

\section{Methodology}

\subsection{Dataset}

The system is tested on two different test set. These are  NTU-RGBD\cite{shahroudy2016ntu} cross-subject test-split and our manually collected test set.

\subsubsection{NTU-RGBD}

It provides RGB, Depth, infrared and 3D Skeleton video collected by three Microsoft Kinect V2; however, we only use skeleton data including 25 joints per person. There are 60 action classes in the dataset and 56880 action samples in total. 

The general way to split this dataset is cross-view or cross-subject. Either we can use the same views or the same subjects for both training and test sets. In this work, we introduce our results just on the cross-subject split. 

Although it is a standard massive benchmark dataset, the data collection system is highly structured and the data is too clean for a real-world scenario. 

\subsubsection{Our Test Set}

We collected a total of 85 skeleton sequences by using our data collection system. The actions are performed by three different people. However, this information is not yet used for any training or testing. There are 19 classes in the dataset which are a subset of NTU-RGBD dataset. The distribution of data can be seen in Table~\MakeUppercase{\romannumeral 1}.  Collecting this small dataset is critical because it allows us to assess the robustness of the modern action recognition algorithms trained on a conventional dataset.

\begin{table}[]
	\centering
	\resizebox{0.34\textwidth}{!}{%
	\begin{tabular}{|c|c|}
		\hline
		\textbf{Action Class}        & \textbf{Number of Video Samples} \\ \hline
		Cheer Up                     & 3                                \\ \hline
		Point to something           & 6                                \\ \hline
		Cross hands in front of body & 3                                \\ \hline
		Throw                        & 6                                \\ \hline
		Wave hand                    & 4                                \\ \hline
		Standing up                  & 6                                \\ \hline
		Sitting down                 & 6                                \\ \hline
		Cough                        & 6                                \\ \hline
		Touch chest                  & 6                                \\ \hline
		Touch head                   & 4                                \\ \hline
		Touch back                   & 3                                \\ \hline
		Vomit                        & 3                                \\ \hline
		Drop                         & 5                                \\ \hline
		Kick                         & 6                                \\ \hline
		Take off glasses             & 3                                \\ \hline
		Wear on glasses              & 4                                \\ \hline
		Eat                          & 4                                \\ \hline
		Drink                        & 4                                \\ \hline
		Read                         & 4                                \\ \hline
		Total                        & 85                               \\ \hline
	\end{tabular}
}
\bigskip
\label{tab:data-stats}
\caption{Action classes and number of samples per each class in our dataset. All samples are stored as ROS bag files which contain the extracted human skeletons.}
\end{table}

\subsection{Framework}

\subsubsection{Hardware}

\begin{table*}[t!]
	\centering
	\begin{tabular}{|l|c|c|}
		\hline
		\multicolumn{1}{|c|}{\textbf{Model}}            & \textbf{NTU-RGBD Cross-Subject Test Accuracy} & \textbf{Our Test set Accuracy} \\ \hline
		Baseline                                        & 74\%                                         & 20.98\%                           \\ \hline
		Baseline + noise                                & 75.17\%                                          & 23.45\%                           \\ \hline
		Baseline + augmentation                         & 75.3\%                                           & 16.04\%                           \\ \hline
		Baseline + augmentation+ noise                  & 74.3\%                                           & 25.92\%                           \\ \hline
		Baseline + normalization                        & 74.6\%                                           & 40.71\%                           \\ \hline
		Baseline + normalization + noise                & 71.6\%                                           & 37.03\%                           \\ \hline
		Baseline + normalization + augmentation         & 70\%                                             & 49.38\%                           \\ \hline
		Baseline + normalization + augmentation + noise & 68.9\%                                           & 46.91\%                           \\ \hline
	\end{tabular}
	\label{tab:results}
	\bigskip
	\caption{Results on two different dataset. The accuracies are lower when the system is applied to another dataset. Even though the augmentation, normalization and noise additions are not required for higher accuracies on the NTU-RGBD test set, they are crucial for transferability.  }
\end{table*}

The training hardware contains NVIDIA GTX1080ti and the test hardware contains NVIDIA GTX1060 which is a laptop. The RGBD camera used in our experiment is Orbbec Astra Pro, as a standard replacement for Microsoft Kinect.  

The test hardware is intentionally kept simple since it is not possible to load so much computation power on a service robot for the reasons below: 

\begin{itemize}
	\item Price,
	\item Heat dissipation, 
	\item Space limitation, 
	\item Power consumption.
\end{itemize}

\subsubsection{Software}

The whole system is implemented on Robot Operating System(ROS)\cite{quigley2009ros} so that it can be directly transferable to any robot. The overall structure of the system is depicted in Figure~\ref{fig:liveaction}. First, the camera collects RGBD data and publishes it. This data is processed by the Body Tracker module to extract the skeleton joints from each frame. Then, these joints are formatted and packed so that they can be used for action recognizer. This module packs three seconds of data which is optimal for our case because if it packs less data, the information may be low and if it packs more than three seconds of data, then the latency of the system would be too high. However, this setting can be adapted according to the scenario. Finally, the Action Recognizer processes the data and publishes the action labels. 

Transferability also requires modularity, since it must be possible to change only one module while deploying the system to different robots or media. For instance, if body tracker is changed in our system, only the joint formatter should be adapted.

The human skeleton detection is beyond the scope of this work; that's why we use Orbbec Astra SDK provided with the camera for this purpose. It tracks the 19 skeleton points of human and gives 3D coordinates of each point. 

The action recognition module is a convolutional neural network based on a previous work\cite{kim2017interpretable} because it is a fast and light network which makes it suitable for robotics applications. Also, their training routine and augmentation are used. In general, the effect of following operations are considered: 

\begin{itemize}
\item \textbf{Removal of the joints} is done to match the joints supported by dataset and our skeleton extractor. Also, some unstable joints are removed such as hands, since their detection rate is lower than the others.    

\item \textbf{Adding Noise} is done via adding zero mean $\sigma$ variance Gaussian noise to the raw skeleton data during the training. This helps emulating the variations between the actions of different people. Our claim is that it helps making the system more robust in real world scenarios.   

\item \textbf{Augmentation} is applied by shifting the skeleton sequence by a random number of frames in time. Also, the random cropping is applied to simulate the missing data in real world scenarios.  

\item \textbf{Normalization} is done via scaling the whole skeleton by a constant value. Additionally, all skeletons are rotated so that the line between shoulders is aligned for all skeletons in the dataset. 

\end{itemize}

\section{Results}

The all models are trained on the cross-subject training split of NTU-RGBD dataset. 

The results are presented in Table~\MakeUppercase{\romannumeral 2}. It is clear that there is a big gap between accuracies on two different dataset. The normalization is necessary for a practical application, because the size of the people changes a lot with the changing distance to cameras. Additive noise during training also improves the accuracy; however, it is not necessary to achieve the best results. The network trained with augmented and normalized training set yields the best results. 

\section{Discussion and Conclusion}

In this work, we show the challenges of designing a Skeleton-based Action Recognition system for real world scenarios. This work is important especially for service robotics domain, since the humans can be anywhere in the scene and some of the joints may not be visible to the robot's camera. Therefore, the augmentation and the normalization become important. The results show that transferability is still an issue for these models. 

In the further studies, our main focus will be transferring action recognition algorithms on a real robot in order to use it for a real scenario.  

\section{Acknowledgment}

This work has received funding from the European Unions
Horizon  2020  research  and  innovation  programme  under
the Marie Skodowska-Curie grant agreement No 721619 for
the  SOCRATES  project. 

\bibliography{ref.bib}
\bibliographystyle{plain}

\end{document}